\documentclass[preprint, 5p]{elsarticle}
\graphicspath{ {./figures/} }
\usepackage{hyperref}
\usepackage{float}
\usepackage{verbatim} %comments
\usepackage{apalike}
\usepackage{amsmath}
\restylefloat{figure}
\restylefloat{table}
\usepackage{booktabs}
\usepackage{lipsum}  
\usepackage{eurosym} % para el euro
\usepackage{lineno}
\usepackage{xcolor}

\journal{Journal of Environmental Management}

\biboptions{authoryear}

\begin{document}
\begin{frontmatter}

\title{Integrated Water Resource Management in the Segura Hydrographic Basin: an Artificial Intelligence Approach}

\author[vicom,ehu]{Urtzi Otamendi \corref{cor1}}
\ead{uotamendi@vicomtech.org}
\author[vicom]{Mikel Maiza}
\author[vicom]{Igor G. Olaizola}
\author[ehu]{Basilio Sierra}
\author[vicom]{Markel Flores}
\author[vicom]{Marco Quartulli}
\cortext[cor1]{Corresponding author.}
\address[vicom]{Vicomtech Foundation, Basque Research and Technology Alliance (BRTA), Donostia-San Sebastián 20009, Spain}
\address[ehu]{Department of Computer Sciences and Artificial Intelligence, University of the Basque Country (UPV/EHU), Donostia-San Sebastián 20018, Spain}

\begin{abstract}
Managing resources effectively in uncertain demand, variable availability, and complex governance policies is a significant challenge. This paper presents a paradigmatic framework for addressing these issues in water management scenarios by integrating advanced physical modelling, remote sensing techniques, and Artificial Intelligence algorithms. The proposed approach accurately predicts water availability, estimates demand, and optimizes resource allocation on both short- and long-term basis, combining a comprehensive hydrological model, agronomic crop models for precise demand estimation, and Mixed-Integer Linear Programming for efficient resource distribution. In the study case of the Segura Hydrographic Basin, the approach successfully allocated approximately 642 million cubic meters ($hm^3$) of water over six months, minimizing the deficit to 9.7\% of the total estimated demand. The methodology demonstrated significant environmental benefits, reducing CO2 emissions while optimizing resource distribution. This robust solution supports informed decision-making processes, ensuring sustainable water management across diverse contexts. The generalizability of this approach allows its adaptation to other basins, contributing to improved governance and policy implementation on a broader scale. Ultimately, the methodology has been validated and integrated into the operational water management practices in the Segura Hydrographic Basin in Spain.
\end{abstract}


\end{frontmatter}

%\linenumbers

  \section{Introduction}
\label{introduction}

%The importance of water management

Efficient and sustainable water resource management constitutes a critical challenge due to increasing demand and environmental constraints. With approximately 4 billion people confronting water scarcity annually \citep{mekonnen2016four}, threats like pollution and climate change-induced droughts and floods are increasing. Effectively managing these challenges requires a holistic approach and integrated solutions for water management.

The Segura Hydrographic Basin, situated in southeastern Spain, is a semi-arid region characterized by a favourable climate and a scarcity of rainfall \citep{lopez1998vegetation, swyngedouw2014not}. These distinctive aspects, combined with the geological characteristics and soil richness, make it conducive to agricultural activity \citep{puy2013genesis, aguillo2002agua}. Approximately 87\% of resources are exclusively dedicated to agrarian practices \citep{CHS}. However, over the past three decades, water demand has consistently exceeded availability, resulting in a persistent water deficit aggravated by intensive exploitation and recurring droughts \citep{pulido2022impact, garcia2015assessing}.

\begin{figure*}[ht]
    \centering
    \includegraphics[width=\linewidth]{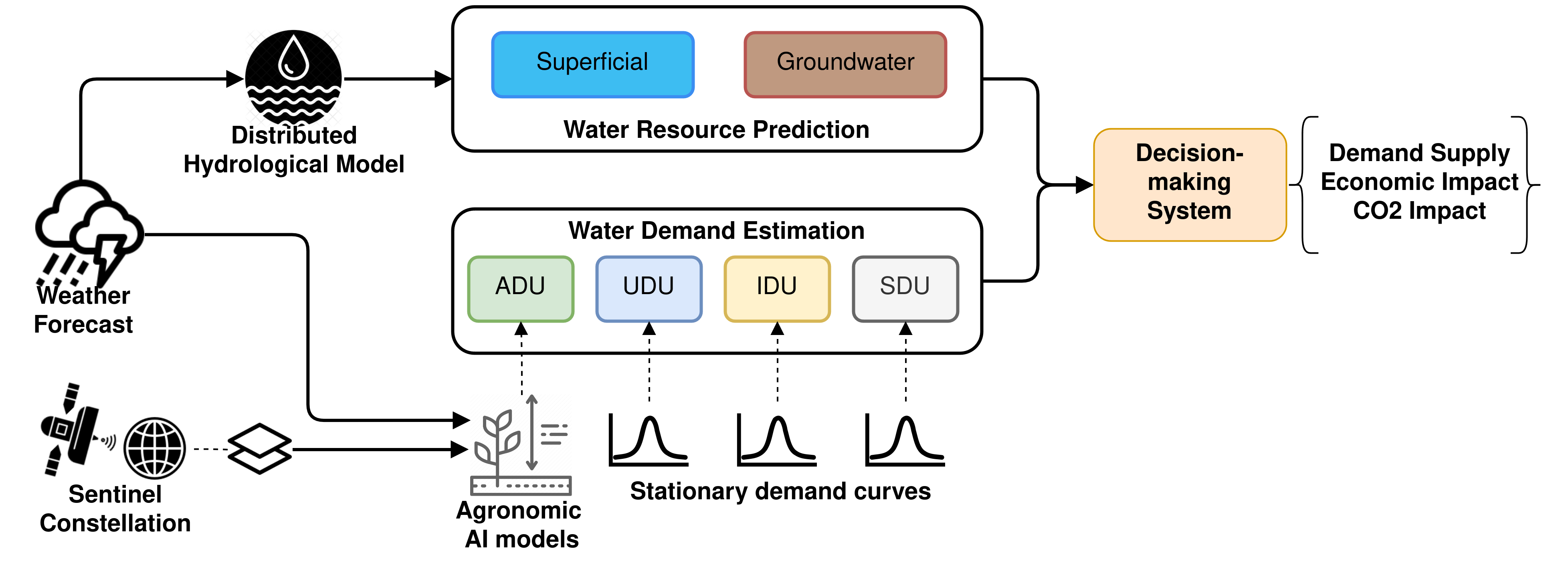}
    \caption{Overview of the proposed Integrated Water Resources Management pipeline in the Segura Hydrographic Basin.}
    \label{fig: pipeline}
\end{figure*}

In response to the intrinsic environmental constraints of the region, the escalating demand, and water pollution, optimum water resource management has emerged as an important necessity \citep{10.2166/wrd.2013.044}. In the last decades, several studies and works have conducted a shift away from sectoral and disjointed management toward the adoption of Integrated Water Resources Management (IWRM) \citep{ALDAYA2019755}, facilitating a broader analysis that encompasses the interconnection of social, economic, and governance factors \citep{GRINDLAY2011242}. IWRM is inherently complex, particularly when considering the role of humans within these systems. It is crucial to highlight clear assumptions regarding the limitations imposed by human factors and their significant influence on the efficacy of IWRM.

%Previous works in the literature and actual work in the area
% Importance of polution
% Economical aspect

In resource management and decision-making, it is imperative to incorporate an understanding of the dynamic environmental conditions \citep{trunk2020current, xiang2021urban}. Within this application context, the scenario is subject to active changes due to environmental factors, such as climate variability or seasonal fluctuations in resource use. Therefore, precise monitoring and forecasting of these changes are crucial \citep{chen2021reinforcement, wang2009irrigation, hamlet2000long}. Further, water capacities and resources exhibit variability, as do the associated demands \citep{CHS}. Considering both aspects and leveraging the capacity to estimate and predict future conditions, decision-making capacities are enhanced, facilitating optimal and efficient management of water resources \citep{MURATOGLU2022157396}.

%% Knowledge gaps
Current Integrated Water Resources Management systems use simulation models and historical data for decision-making. Most of these techniques lack the use of forward-looking data, contemplating only a static environment, even though we are dealing with a dynamic scenario such as the hydrological one. The use of artificial intelligence prediction techniques has helped to improve the performance of other decision-making domains. In this work, we propose as a novelty the use of AI models, machine learning, and generative AI to improve the monitoring of the environment and the consequent decision-making.

This paper presents an approach for Integrated Water Resources Management in the Segura Hydrographic Basin (refer to Figure \ref{fig: pipeline}). The work focuses on developing an intelligent decision-making system to facilitate efficient and sustainable water resource management in the region. This methodology offers a comprehensive and integrated pipeline using advanced techniques to forecast available resources and estimate demands across the basin's expanse. The proposed solution optimizes the resources-to-demands distribution of water on both a daily and long-term (6-7 months) basis. The daily or short-term simulation takes 15 days, defined by the governance processes in the basin. The novel approach considers the diverse water demand fluctuations (urban, agricultural, industrial, touristic, and wetlands) and important KPIs associated with the different uses of water, such as environmental impact, measured in CO2 emissions, and economic impact.

This decision-making system encompasses multiple stages throughout the pipeline:

\begin{enumerate}
    \item \textbf{Weather Forecasting}: combines data from atmospheric forecasting services and historical data from local weather stations. This information is processed via a Weather Research and Forecasting (WRF) system to extract crucial variables, subsequently adjusting them to align with Segura Basin's specific features.
    \item \textbf{Distributed Hydrological Model (DHM)}: it entails the simulation of water flow, storage and dynamic interaction of interconnected hydrological components using historical and real-time sensor data combined with weather forecasts.
    \item \textbf{Agronomic Modeling}: it is performed by leveraging historical earth observation (EO) and weather forecasting data using Artificial Intelligence (AI) techniques. This stage includes developing agronomic crop models to predict irrigation demands and biomass production.
    \item \textbf{Decision-making system}: it makes use of a Complementary Hydrological Model (CHM) representing the Segura Hydrographic Basin to simulate the effects of the decisions in the water availability. The algorithm determines the optimal actions to supply the predicted demand with estimated and simulated resources while considering CO2 emissions and economic impact.
\end{enumerate}

As a result of the research work, the proposed decision-making approach is validated and integrated within a real operational framework. The algorithms implemented in this system are connected with real-time sensing systems, thus providing the capability of making predictions and estimations for the monitoring data.

The remainder of the article is structured as follows: Section \ref{sec: problem} formally describes the hydrological structure of the Segura Hydrographic Basin and the central water management problems. Section \ref{sec: methodology} describes the proposed solution and associated pipeline of services, briefly introducing technological details. Section \ref{sec: Results} illustrates the functioning of the decision-making process in a real scenario, including the description of some exemplary results and their discussion. Section \ref{sec: Integration} presents the integration of the developed approach in a real operational scenario. Finally, Section \ref{sec: conclusions} contains the concluding remarks.

\section{Problem description}
\label{sec: problem}

% Explain the problem in the basin

The Segura Hydrographic Basin is one of Europe's most intricate river systems \citep{grindlay2011atomic}, presenting a critical challenge for Integrated Water Resources Management. In addition to limited water resources, governance complexities increase management difficulties, such as intense competition among water users, especially in agriculture.

The frequency of pollution incidents in water bodies has increased the need for strict regulations on surface water resources \citep{GOMEZGOMEZ2022153128}. As a consequence, reliance on groundwater is significant, supplemented by unconventional sources like transfers from neighbouring river basins, treated wastewater reuse, and desalination plants \citep{10.2166/wrd.2013.044, LAPUENTE201240}. Managing water demand in the basin involves addressing several key factors:

\begin{itemize}
    \item Adjusting water volume demands while considering supply guarantees, quality standards to combat pollution, and economic and environmental impact.
    \item Legal constraints related to water use rights and water transfer regulations. For instance, governance complexity considerably increases when historically determined irrigation rights in agricultural activities do not correspond with the required irrigation demand of the harvest. 
    
\end{itemize}

Moreover, with agriculture dominating water usage (87\%), competition for water use rights intensifies. The basin annually demands an average of $1,700 hm^3$ of water, where $800 hm^3$ are from the basin's natural resources \citep{CHS_2022}. Strategic measures such as promoting reuse, regulating flows, managing aquifer exploitation, and implementing desalination are critical. 

Regarding water quality, the Segura Hydrographic Basin faces several water pollution problems. These incidents, arising from industry, agriculture, and urban areas, considerably degrade water, aggravating the pressure on limited resources. Agricultural activities aggravate pollution due to fertilizers and pesticides, contributing to contamination. Urbanization and industrialization also contribute to pollution, with wastewater discharges containing heavy metals and organic compounds posing risks to ecosystems and human health.

The above issues amplify the challenge of ensuring water quality and sustainability. Managing the imbalance between resources and demand requires comprehensive strategies tackling scarcity and pollution issues. Integrated approaches enclosing pollution prevention, efficient water use and reuse, and ecosystem restoration are critical to mitigating these dual challenges.

\subsection{Available Resources}

The Segura Hydrographic Basin spans an approximate surface area of 19,025 km². Its hydrographic network primarily comprises the Segura River and its branches. The basin has been subdivided into 14 hydraulic zones, taking into account hydrographic, administrative, socioeconomic, and environmental factors \citep{CHS_anejo12}.

\begin{figure}[ht]
    \centering
    \includegraphics[width=\linewidth]{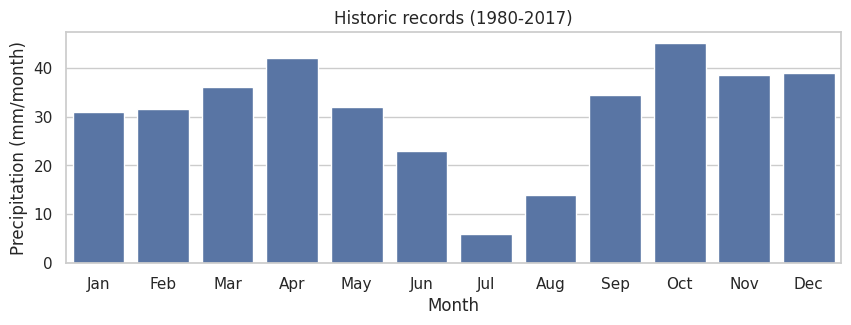}
    \caption{Historical average accumulated monthly precipitation from 1980/81 to 2017/18 \citep{simpa}. }
    \label{fig: rain}
\end{figure}

Precipitation data from 1980/81 to 2017/18 \cite{simpa} reveals that the average annual precipitation in the basin varies. It exhibits seasonal and spatial variations, with wet months occurring mainly in autumn and spring and dry months prevailing in summer (refer to Figure \ref{fig: rain}). Geographically, rainfall is concentrated in mountainous regions, with some areas experiencing annual rainfall exceeding 1,000 mm/year.

Water bodies in the basin are classified into two main groups: surface water and groundwater. Groundwater bodies, defined as aquifers or groups of aquifers managed collectively, are 63 and comprise 242 individual aquifers.

Surface water water bodies are further categorized into rivers, lakes, transitional, and coastal waters based on their nature. These categories may include natural, highly modified, or artificial water bodies. For instance, dams are categorized as modified or artificial lakes  \citep{CHS_anejo12}. Dams and lakes utilized as salt pans also fall under this category. The CHS identifies 1,553 km of significant natural rivers in the basin \citep{CHS_anejo12}.

% \begin{itemize}
%    \item  \textbf{Rivers}: the CHS identifies 1,553 km of significant natural rivers in the basin \citep{CHS_anejo12}. Natural rivers are characterized based on average flow, while channelized rivers are considered highly modified water bodies.

%    \item  \textbf{Lakes}: a water body is classified as a lake if its surface area exceeds 50 ha ($0.5 km^2$) or exceeds 8 ha with a maximum depth greater than 3m. Lakes affected by irrigation or drainage infrastructures or regulated are categorized as highly modified water bodies. Dams and lakes utilized as salt pans also fall under this category.

%    \item  \textbf{Transitional Bodies}: these are surface waters near river mouths, lakes, lagoons, or wetlands that are partially saline due to coastal proximity but significantly influenced by freshwater flows.

%    \item  \textbf{Coastal Waters}: surface waters within one nautical mile offshore from the nearest point on the baseline are classified as coastal waters.
% \end{itemize}

\begin{figure}[ht]
    \centering
    \includegraphics[width=\linewidth]{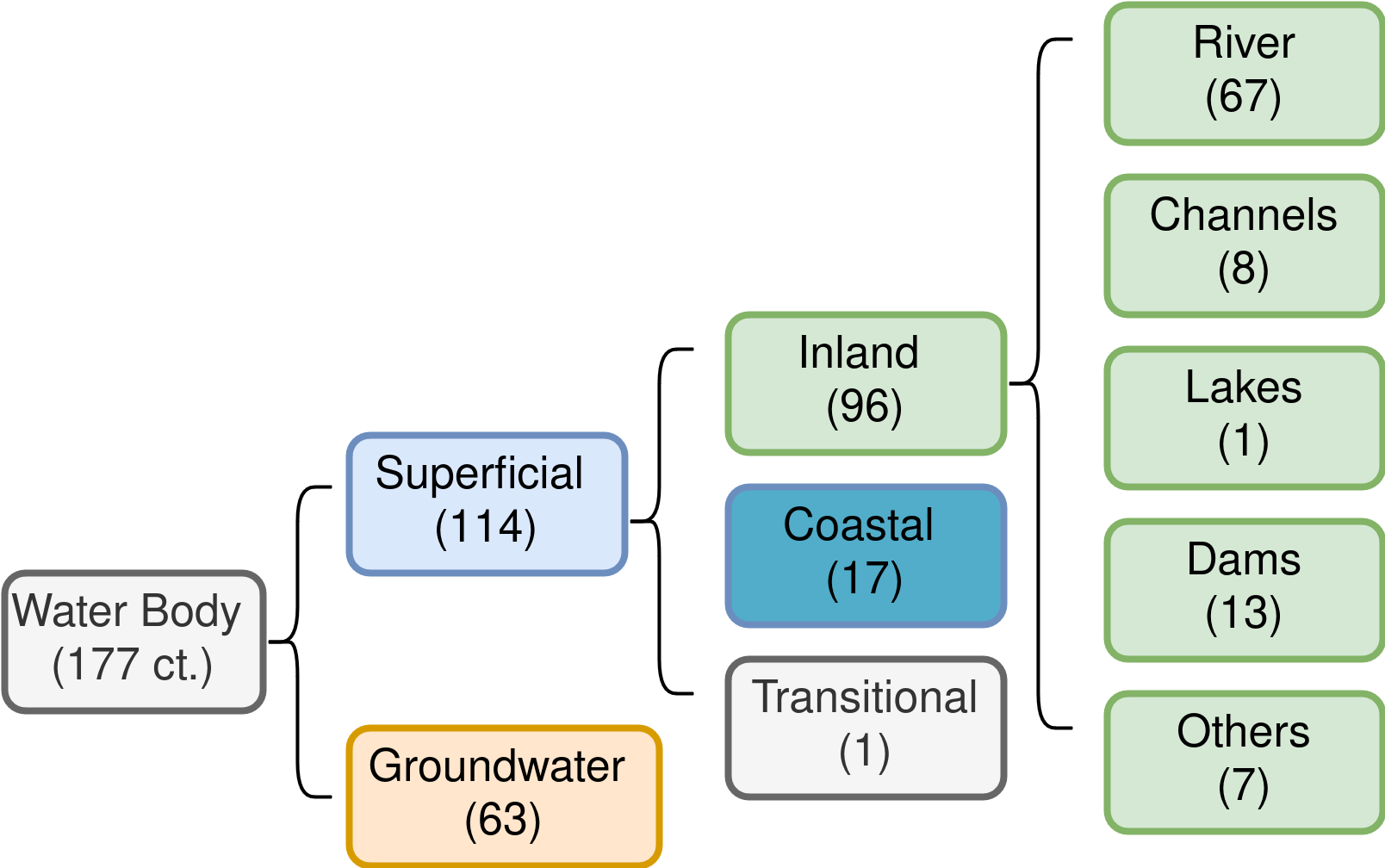}
    \caption{Categorised water bodies in the Segura Hydrographic Basin \citep{CHS_anejo12}.}
    \label{fig: resource_types}
\end{figure}

The basin comprises 177 water bodies, including rivers, lakes, reservoirs, and groundwater sources (refer to Figure \ref{fig: resource_types}).

Determining the flow rates and stored resources in these bodies is complex. For decades, resource monitoring has been conducted to collect historical data and analyse fluctuations \citep{erena2019use, galiano2011monitoring}. These measurements are invaluable for realistically estimating the bodies' water storage. Due to their underground nature, aquifers pose significant challenges in monitoring.

According to a study conducted by the \cite{CHS_anejo02}, the Segura Hydrographic Basin manages an annual natural water supply of $1,410hm^3$, based on historical records from 1980/81 to 2017/18 (refer to Table \ref{tab: resources}). Within this water resource allocation, $764hm^3$ are sourced from water bodies delineated within the basin (refer to Figure \ref{fig: resource_types}). According to this study, $491hm^3$ originate from aquifers (groundwater), while the remaining $273hm^3$ are from surface water resources.

Due to water scarcity within the demarcation, unconventional water sources, such as desalination and water reuse, have been increasingly implemented in recent decades \citep{CHS_2022}. Accordingly, desalination processes produce $302hm^3$ of water, with $223hm^3$ allocated for agricultural purposes and the remainder for urban or industrial use. Regarding water recycling, $142hm^3$ is derived from urban or industrial demand reuse, while $121hm^3$ is obtained from irrigation runoff to surface and groundwater bodies. Additionally, coastal bodies, such as aquifers and wadis that do not discharge into the river, contribute $81hm^3$.

\begin{table}[!ht]
    \centering
    \begin{tabular}{rr}
                 & $hm^3/year$\\
                 \toprule
         Natural sources of the Basin&  764\\
         Inter-community basin transfer&312\\
         Desalinization&  302\\
         Water Recycling&263\\
         Not draining Coastal bodies&81\\
   \midrule
&1722 $hm^3/year$\\
    \end{tabular}
    \caption{ Average resources per year of the Segura Hydrographic Basin \citep{CHS_anejo02}, considering the contributions of other inter-community basins. }
    \label{tab: resources}
\end{table}

Additionally, resources from other inter-community basins amount to $312hm^3$, with $295hm^3$ supplied by the Tajo-Segura aqueduct and the remainder sourced from Negrat\'in.

However, these resource availability are based on historical measurements, introducing a significant source of uncertainty. Weather conditions in the current and preceding years highly influence the state of basin bodies. Prolonged droughts and years of scarce rainfall generate deviations from historical data, emphasising the importance of real-time monitoring and forecasting to characterize the actual condition for water management scenarios.

\subsection{Demand Units}

In the hydrological plan of the basin, water demands encapsulate the various utilization and activities that impact water resources. The Segura Demarcation Hydrological Plan \citep{CHS_anejo3} delineates four primary categories of water usage: (1) population supply, (2) agricultural demand, (3) industrial usage, (4) service usage, such as golf courses, and (5) environmental demand for wetland maintenance. The CHS partitions the demand into demand units to ensure efficient governance of these demands. A demand unit refers to demands of the same category that use common sources and reintegrate their returns in the same geographical area.

Each demand unit has distinct characteristics depending on the usage category. For instance, urban demand units, encompassing domestic, public, and commercial usage, are prioritized over other uses, such as recreational activities like golf courses.

Considering the economic impact of water usage, it is critical to analyse the economic activities reliant on demand units, measure how water usage influences various sectors such as agriculture, industry, and tourism, and identify differences in economic contributions among different usage types and supply costs from each resource category.

\textbf{Urban Demand Units (UDU)} includes domestic use, supply to local and institutional public services, and water service for businesses and industries in the municipal area. The demand for a UDU is predicated on recorded water consumption, including non-recorded sources like losses (leaks), non-measured authorized consumption, or estimated unauthorized consumption. Urban demand is subdivided into 13 units.%, as depicted in Figure \ref{fig: UDUs}.

For urban areas, governmental institutions clearly outline specific legal water allocation criteria. Demand is considered satisfied when the deficit within a month does not exceed 10\% of the corresponding monthly demand and the cumulative deficit does not surpass 8\% of the annual demand over ten consecutive years. 

% \begin{figure}[ht]
%     \centering
%     \includegraphics[width=\linewidth]{figures/udus.png}
%     \caption{Map of the Urban Demand Units (UDU) in the Segura Hydrographic Basin \citep{CHS_2022}.}
%     \label{fig: UDUs}
% \end{figure}

\textbf{Agricultural Demand Units (ADU)} encompass areas where water usage includes agricultural, forestry, and livestock demands. In ADUs, agricultural irrigation constitutes the primary use. Measuring the demand of each ADU is complex due to diverse crops, geolocation, and other variables affecting production. Consequently, agricultural demand emerges as the relationship between theoretical water requirements and overall irrigation efficiency, combining actual irrigation demand with resource allocation efficiency.

\begin{figure}[ht]
    \centering
    \includegraphics[width=\linewidth]{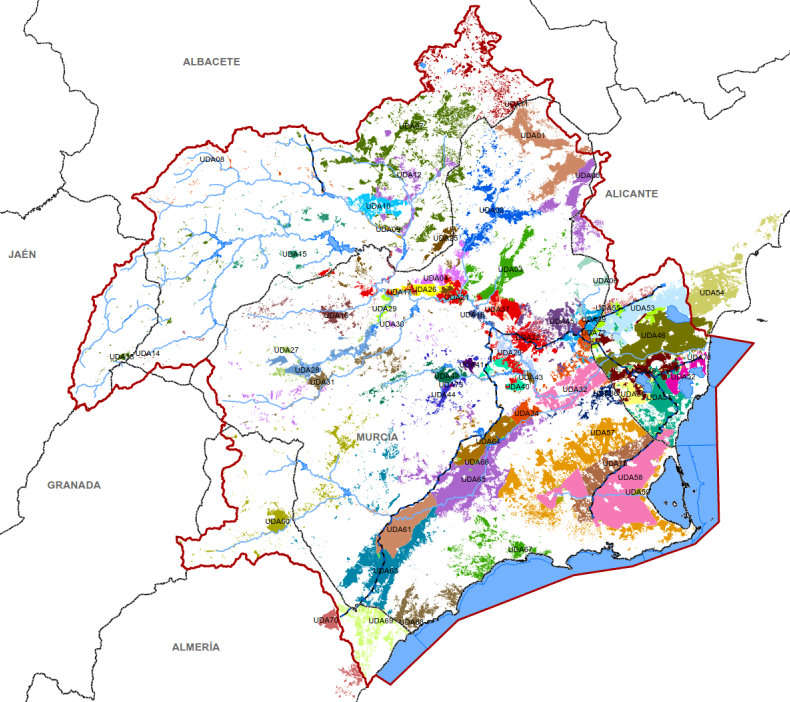}
    \caption{Map of the Agricultural Demand Units (ADU) in the Segura Hydrographic Basin \citep{CHS_2022}.}
    \label{fig: ADUs}
\end{figure}

The computation of irrigation demand integrates various factors from the total area within the digitalized surface of irrigation rights, including unproductive areas, crop rotation, and unproductive section deductions. Additionally, annual remote sensing campaigns are used to recalculate actual irrigated surfaces \citep{CHS_anejo3}. Within this context, water demand is estimated by determining the necessary water per unit of irrigated area for optimal crop production, considering crop type, climatic attributes, and hydrological factors.

In the CHS study, the effective irrigation area of ADUs spans $448.5ha$, divided among 62 ADUs as depicted in Figure \ref{fig: ADUs}.

\textbf{Industrial Demand Units (IDU)} are the areas primarily dedicated to manufacturing industry activities, excluding extractive, energy, and construction activities. Industrial water supply may be sourced from urban supply networks (connected) or proprietary sources (non-connected). These demands are estimated from industrial subsectors and area considerations. 7 IDUs are identified as non-connected in the basin, while connected industrial areas are subsumed within the corresponding UDU.

\begin{table}[!ht]
    \centering
    \begin{tabular}{rrr}
   
            & $hm^3$ &\%\\
        \toprule
        ADU &  1,476.3 & 87\\
         UDU &  200.9 & 11.84\\
         SDU & 11.2 & 0.66\\
         IDU & 8.5 & 0.5\\
          \midrule
        &1,696.9 &\\
    \end{tabular}
    \caption{ Estimated Water demands of the Segura Hydrographic Basin in 2021 \citep{CHS_2022}. }
    \label{tab: demands}
\end{table}

\textbf{Service Demand Units (SDU)} represent a demand for energy production and service and recreational uses not connected to municipal supply networks. Regarding energy production, the demands include two solar thermal power plants, three thermal power plants and other hydroelectric uses. Recreational and tourism services are mainly focused on the maintenance of golf courses. The golf course irrigation area is estimated by remote sensing at $1,400ha$. For each ha of the golf course, it is assumed irrigation of $8,000 m^3/ha/year$. Within the Segura Hydrographic Basin, 10 SDUs are identified.

\textbf{Wetland Demand Units (WDU)} represents the environmental demand for wetland maintenance, which is additional to the environmental maintenance flows. Wetlands in the basin encompass the following elements:  crypto-wetlands, coastal lagoons or salt marshes, inland salt marshes and lagoons \citep{CHS_anejo3}. For water management, a consumptive environmental demand is defined for the maintenance of these wetlands ($31.67hm^3$ per year). However, these demands are not included within the basin's water supply management, as it is considered an additional system constraint.

According to the study performed by \cite{CHS_anejo3}, the total demand in 2021 is estimated to be $1,696.9hm^3$ (refer to table \ref{tab: demands}). The table shows the dominance of agricultural demand in water usage with 87\% of the demand, underscoring the importance of targeted water management strategies in this sector.

\section{Material and methods}
\label{sec: methodology}

This section provides an overview of the technological framework developed and the water resource management strategy proposed for the Segura Hydrographic Basin, addressing the challenges posed by government complexities and water scarcity issues inherent to the region. Given the hydrological dynamics, containing aquifers and constrained by limited precipitation, accurately monitoring current water resources and predicting future availability is a challenging task. 

Estimating agricultural demand is particularly challenging due to conflicts between established irrigation rights and actual irrigation requirements \citep{RUPEREZMORENO201767}. Thus, this paper introduces a management pipeline integrating advanced techniques to comprehensively characterise the dynamics for facilitating efficient and sustainable water management procedures.

This pipeline comprises interconnected modules (refer to Figure \ref{fig: pipeline}), leveraging one stage's predictive outcomes as inputs for subsequent algorithms to generate further predictions. While the underlying algorithms' intricacies are beyond the scope of this article, the research work focuses on presenting the workflow of the developed framework and providing an overview orchestration of the artificial intelligence techniques employed to characterise the basin.

This section comprises three primary parts: water resource prediction, water demand estimation, and water management optimisation. Each component is crucial in the holistic approach to water resource management.

\subsection{Resource Prediction}
\label{sub: resource_pred}

In this phase, the objective is to predict the availability of water resources in the short- (15 days) and long-term (6 months), leveraging a combination of historical data, real-time measurements, and advanced AI techniques. The weather forecasting and Distributed Hydrological Model (DHM) modules are used for resource prediction.

\subsubsection{Weather forecasting}

The weather forecasting module generates short- and long-term forecasts by combining various prediction systems. For short-term forecasts spanning 15 days, the following public models are utilized: (1) High-Resolution Model (HRS) from the European Centre for Medium-Range Weather Forecasts (ECMWF)\footnote{ECMWF, \url{https://www.ecmwf.int/}}, (2) \textit{Icosahedral Nonhydrostatic} model from Deutscher Wetterdienst\footnote{Deutscher Wetterdienst, \url{https://www.dwd.de}}, (3) Global Forecast System model from National Centers for Environmental Prediction\footnote{National Centers for Environmental Prediction, \url{https://www.weather.gov/ncep/}} and (4) \textit{Action de Recherche Petite Echelle Grande Echelle} (ARPEGE) from Météo-France\footnote{Météo-France, \url{https://meteofrance.com/}}.

\begin{figure}[ht]
    \centering
    \includegraphics[width=\linewidth]{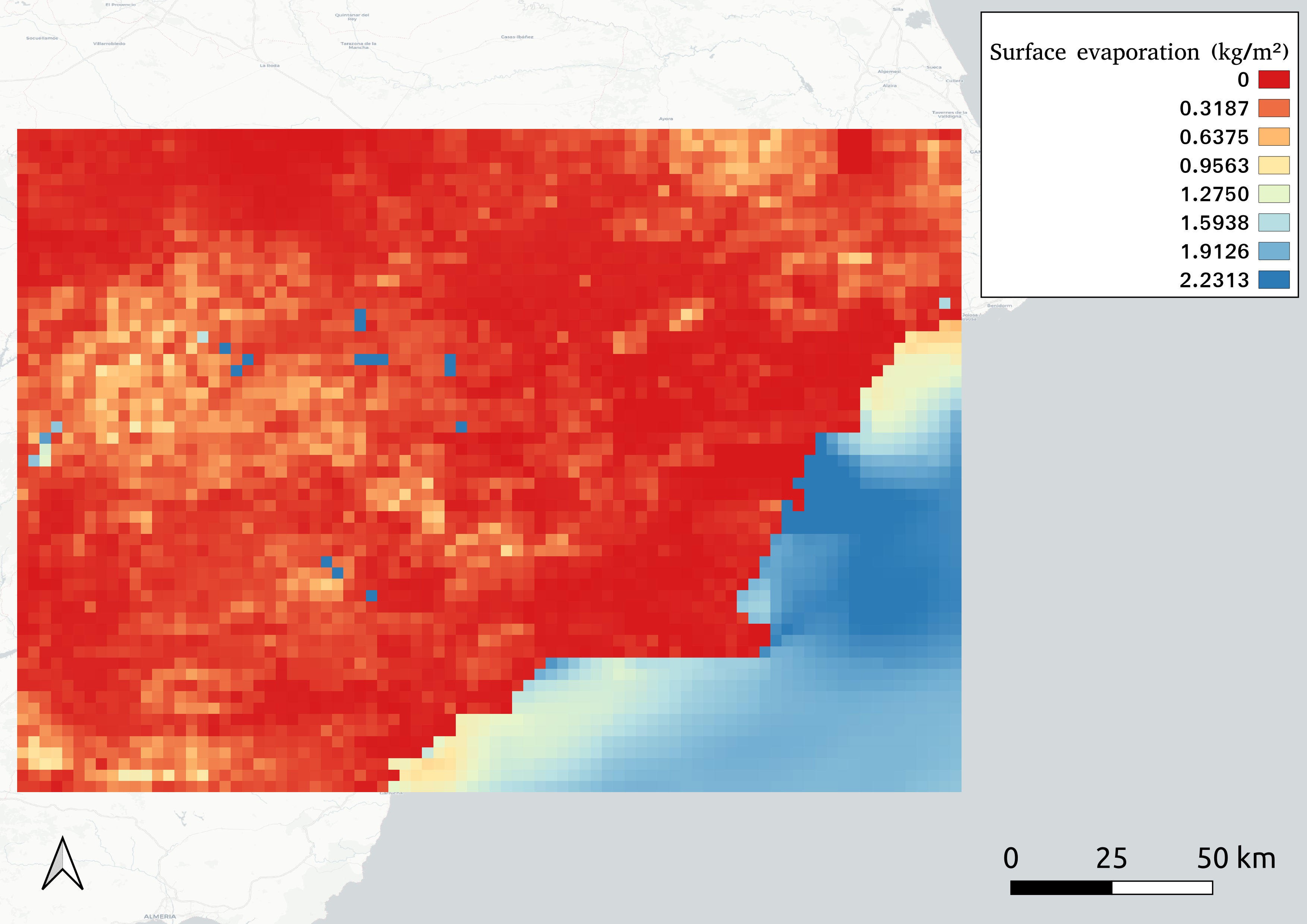}
    \caption{Accumulated surface evapotranspiration forecast example in the Segura Hydrographic Basin.}
    \label{fig: temp}
\end{figure}

In the forecasting process, an initial simulation is conducted using the Weather Research and Forecasting (WRF) technique, leveraging historical records from AEMET\footnote{Agencia Estatal de Meteorología \url{https://www.aemet.es/}} and local weather stations to generate high-resolution predictions in a rasterized grid format. Subsequently, mathematical and statistical procedures are applied to interpolate the required variables using the four forecasting models (refer to Table \ref{tab: forecast_variables}). Finally, a Machine Learning (ML) approach integrates the grid with the interpolated product, enhancing precipitation predictions and yielding an improved forecast product (refer to Figure \ref{fig: temp}).

\begin{table}[htbp]
  \centering
  \caption{The main variables inferred in the weather forecast module.}
  \label{tab: forecast_variables}
    \begin{tabular}{rr}
    
     Variable &  Unit \\
     \toprule
     Accumulated Precipitation & $kg/m^2$ \\
     Mean sea level pressure & $Pa$ \\
     Accumulated surface evaporation & $kg/m^{2}$\\
     Downward short wave flux at ground surface & $W/m^{2}$  \\
     2 metre surface temperature & $K$ \\
     10 metre U wind component &  $m/s$ \\
     10 metre V wind component & $m/s$ \\
     Surface solar radiation downwards & $J/m^{2}$ \\
    \midrule
    \end{tabular}
\end{table}

For the long-term forecasts spanning six months, the following models are utilized: (1) High-Resolution Model (HRS) from ECMWF, (2) Deutscher Wetterdienst (DWD) long-term model, (3) UK Met Office\footnote{UK Met Office\url{https://www.metoffice.gov.uk/}} model and (4) Météo-France. These models offer a spatial resolution of one degree per pixel and a temporal resolution of 24 hours. Similarly to the short-term predictions, the outcomes from these models are integrated with a long-term WRF simulation. These predictions contain the variables outlined in table \ref{tab: forecast_variables}.

\subsubsection{Distributed Hydrological Model}

Using the MIKE SHE\footnote{Dansk Hydraulisk Institut (DHI), \url{https://www.mikepoweredbydhi.com/products/mike-she}} tool, a Distributed Hydrological Model (DHM) is developed based on physical processes implementation and model calibration. This model provides an advanced simulation of the water bodies building networks and estimating surface and groundwater flow. However, the details of this DHM implementation are outside the scope of this paper.

The model is initially developed and calibrated using historical records from 2015 to 2021. Subsequently, real-time monitoring data is incorporated into the online prediction system. The model simulates interconnected components' flow, storage, and dynamic interaction, representing surface flows, unsaturated soil zone flows, groundwater, and rivers.

% \begin{figure}[ht]
%     \centering
%     \includegraphics[width=0.7\linewidth]{figures/mikeshe.png}
%     \caption{River and canal network from the Segura Hydrographic Basin included in the DHM.}
%     \label{fig: mikeshe_model}
% \end{figure}
\begin{figure}[ht]
    \centering
    \includegraphics[width=\linewidth]{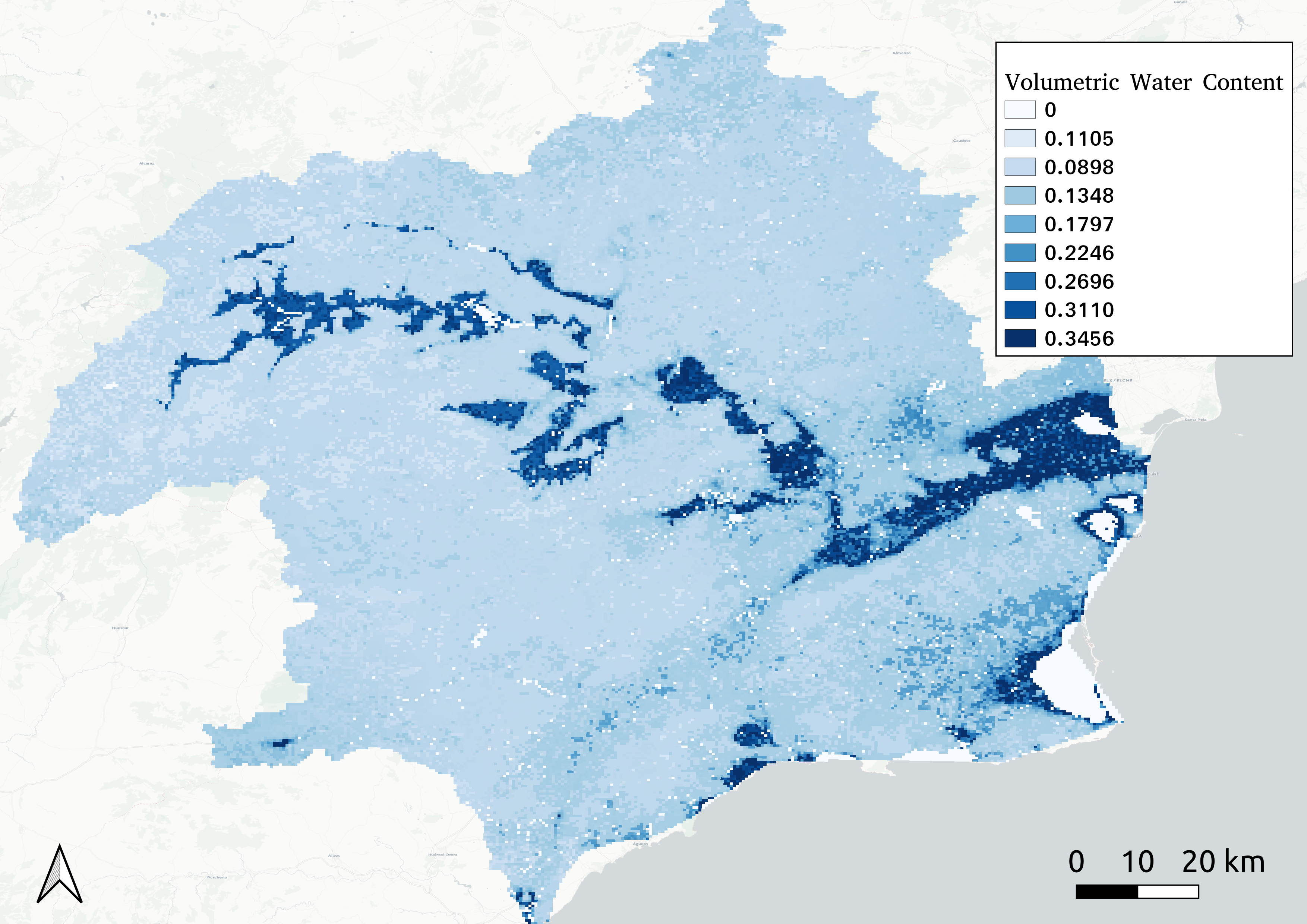}
    \caption{Example of the prediction of average water content in the rootzone generated by the DHM.}
    \label{fig: volumetric}
\end{figure}

The model integrates various variables to establish the hydraulic network and internal dynamics and incorporates weather forecasts from the prior pipeline step, focusing on important predicted variables such as accumulated precipitation and potential evapotranspiration. This information is extracted at a resolution of 3 $km^2$ per pixel, enabling detailed analysis and prediction of hydrological processes. Additionally, topographical data defines the drainage flow in surface bodies, facilitating accurate modelling of water movement within the basin (refer to Figure \ref{fig: volumetric}). Furthermore, soil usage data is essential for delineating urban, industrial, and agricultural areas.

In this predictive process, it is essential to consider the water needs from various demand units and the corresponding resource extractions. For this purpose, irrigation rights and historical trends are applied to estimate water usage demands and returns. 

The hydrological model generates multiple significant outcomes, including:
\begin{itemize}
    \item Integration of real-time data with model-based predictions to enhance accuracy.
    \item Monitoring of storage changes for both surface and groundwater resources.
    \item Assessment of the volume of stored resources.
    \item Time series data to forecast future water availability against established operational thresholds.
    \item Allocation of water resources from diverse sources, encompassing surface and groundwater runoff.
\end{itemize}

The hydrological simulations provide short-term (15 days) and long-term (6 months) predictions, assisting in understanding water dynamics, availability, and optimizing allocation.

\subsection{Demand estimation}
\label{sub: demand_estim}

Estimating water usage is a critical aspect of basin management. Its complexity originates from diverse uses and consumption implications; generating ad-hoc modelling for each demand unit is nearly impractical. Given that agricultural activities (ADU) use 87\% of total demand (refer to Table \ref{tab: demands}), this framework prioritizes irrigation demand prediction. Differences between irrigation rights and actual needs lead to conflicts and waste of water. Inter-annual demand curves are defined based on historical records for other demand units, such as UDU, UDI, and SDU.

\subsubsection{Agricultural demand estimation}

Agronomic models of representative crops are developed to measure the agricultural water demand in the basin. Given the expansive nature of the land areas under analysis, remote sensing techniques emerge as the most efficient approach. Thus, state-of-the-art deep learning techniques are employed to estimate crop water requirements, leveraging satellite multispectral imagery with weather forecasts. The algorithms employed in this agricultural model fall outside the scope of this article. This model is derived from the methodologies presented in prior research, which are comprehensively detailed in three published papers by \cite{guzinski2020modelling, guzinski2021utility, guzinski2023improving}. 

The input data for this module encompasses a range of sources:

\begin{itemize}
    \item Synthetic Aperture Radar data from the Sentinel-1 satellite from the Copernicus Programme\footnote{Copernicus Programme \url{https://sentiwiki.copernicus.eu/web/copernicus-programme}}.
    \item High-resolution optical data from the Sentinel-2 satellite.
    \item Optical and thermal data from the Sentinel-3 satellite.
    \item Phenological data extracted from the Copernicus Land Monitoring System\footnote{Copernicus Land Monitoring Service (CLMS) \url{https://land.copernicus.eu/}}.
    \item Rasterized weather forecasts from the pipeline.
\end{itemize}

\begin{figure}[ht]
    \centering
    \includegraphics[width=\linewidth]{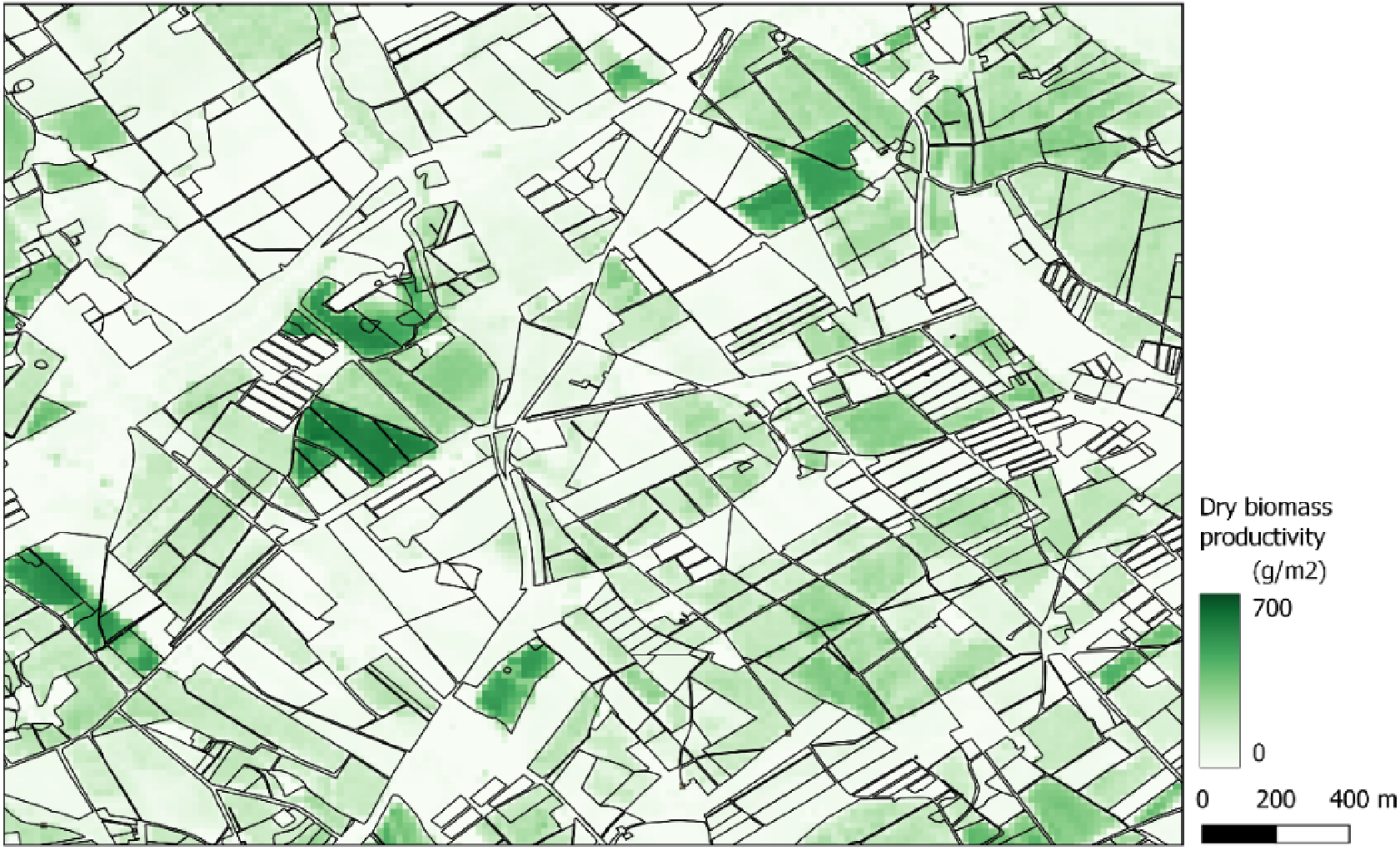}
    \caption{Illustrative example of dry biomass productivity map computed in the crop growth modelling.}
    \label{fig: biomas}
\end{figure}

Utilizing remote sensing data, the leaf area index of plants is computed, enabling the measurement of plant transpiration. Subsequently, the AquaCrop model \citep{raes2009aquacrop} is employed to estimate dry biomass (refer to Figure \ref{fig: biomas}). Following various studies \citep{li2021understanding, nielsen2015cover}, a linear relationship correlates biomass production and water consumption of a given species. These valuable estimations serve for crop growth modelling.

\begin{figure}[ht]
    \centering
    \includegraphics[width=\linewidth]{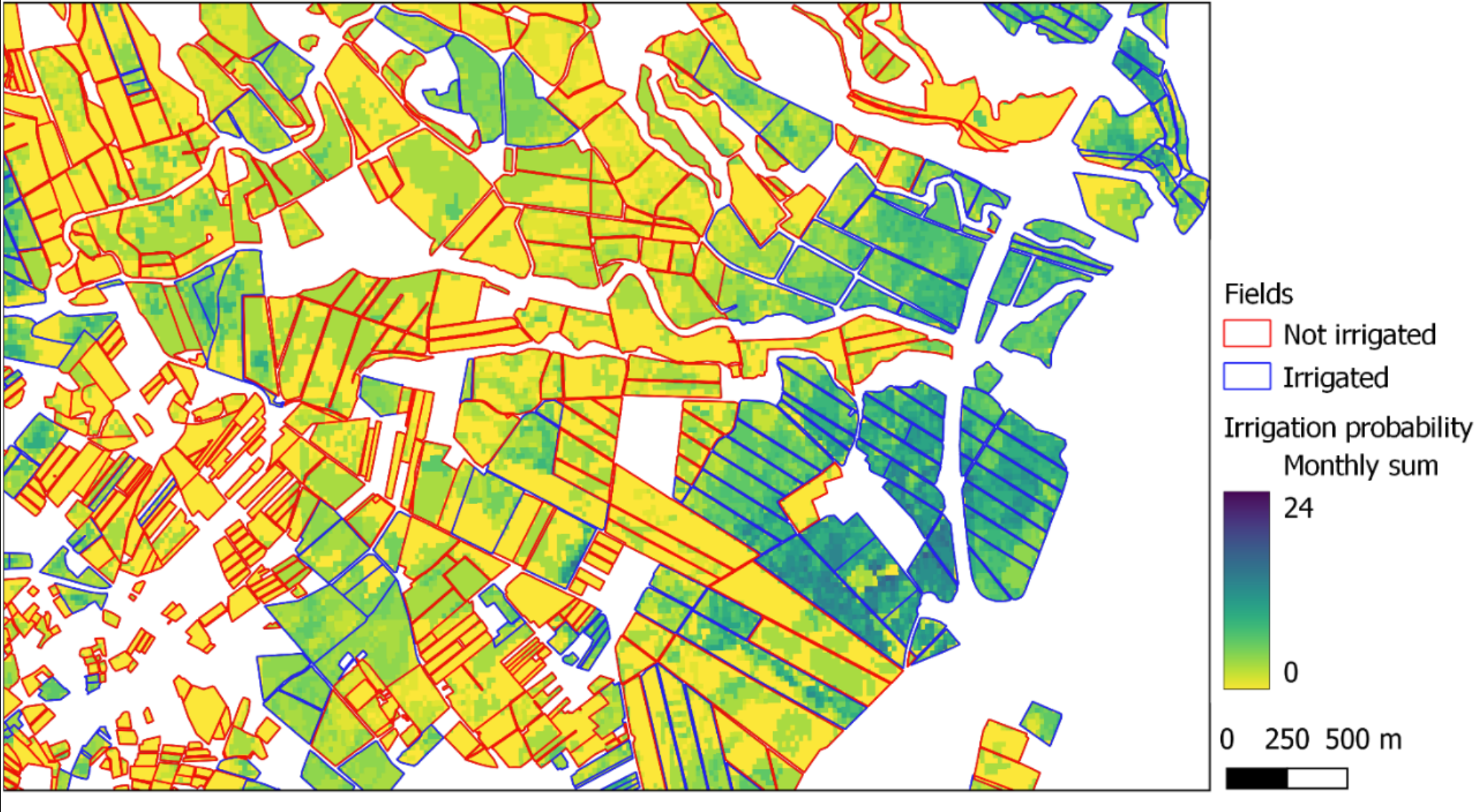}
    \caption{Crop irrigation classification and the probability of irrigation illustrative example.}
    \label{fig: irrigation}
\end{figure}

Integrating remote sensing data with evapotranspiration values enables the identification of recently irrigated areas. Using historical observations facilitates the detection of irrigation activities and provides insights into land usage. Moreover, utilizing inferred historical irrigation data enables the estimation of irrigation probability using weather forecasts and the period of the year (refer to Figure \ref{fig: irrigation}).

Modelling and comparing actual and potential evapotranspiration is instrumental in estimating the crop water deficit. Evapotranspiration, typically derived from an energy perspective using satellite data, encompasses net radiation, ground heat flux, sensible heat flux, and latent heat flux \citep{liou2014evapotranspiration, aryalekshmi2021analysis}. Sentinel-2 and Sentinel-3 satellites facilitate the collection of this vital information.

Thus, the irrigation demands of Agricultural Demand Units (ADUs) are determined by modelling actual evapotranspiration, crop growth dynamics, and historical irrigation probabilities.

\subsection{Decision-making system}
\label{sub: Decision-making}

The estimations mentioned above are pivotal in the last phase of the water management decision-making pipeline. By incorporating future estimates of income (water resources) and demands, the decision-making system achieves a more realistic depiction.

\begin{figure}[ht]
    \centering
    \includegraphics[width=\linewidth]{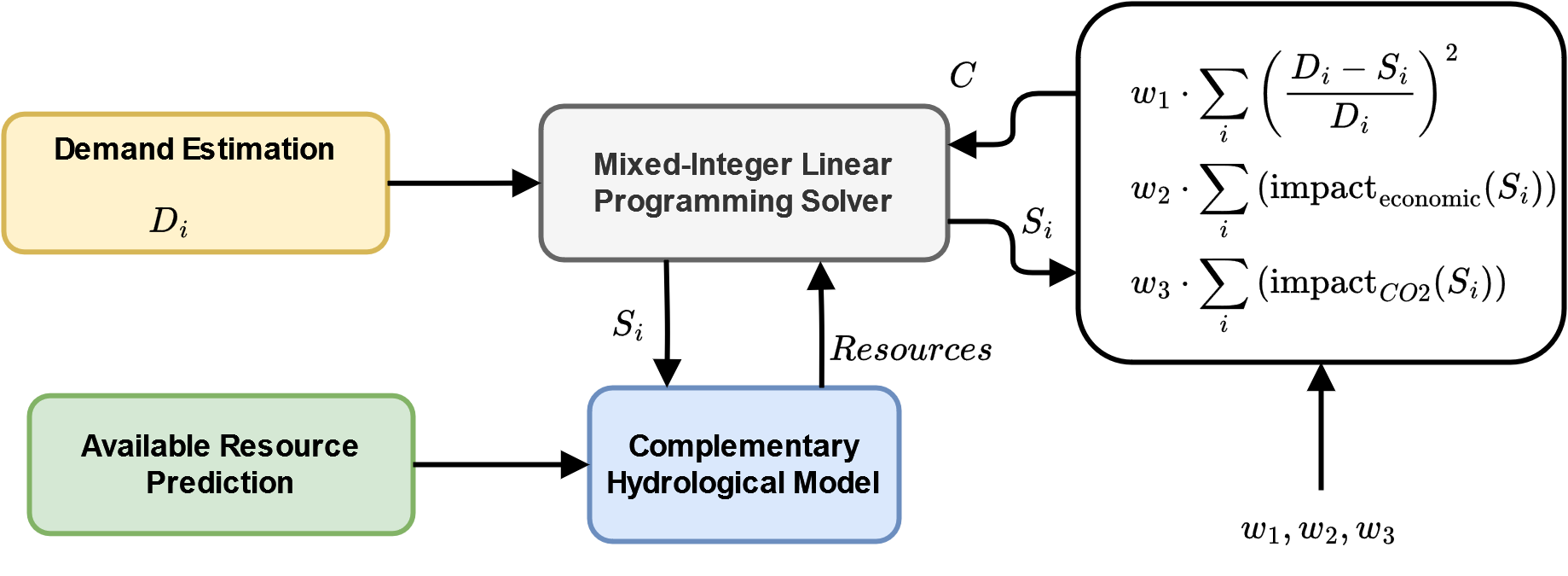}
    \caption{Overview diagram of the decision-making system, where $C$ represents the cost function and $S_i$ represents the water supply to the unit $i$ in the decision-making process.}
    \label{fig: decision-making}
\end{figure}

Decision-making involves selecting optimal actions based on scenario input data, hydrological and mathematical system modelling, and a cost function to evaluate performance (refer to Figure \ref{fig: decision-making}). Previous sections (\ref{sub: demand_estim} and \ref{sub: resource_pred}) outline the generation of input data. This section details system modelling and the associated cost function.

The system modelling operates a Complementary Hydrological Model (CHM), with decision-making variables representing flow management or control within the Segura Hydrographic Basin. This model reproduces hydrological dynamics similarly to the DHM, offering faster simulation but slightly reduced accuracy. The cost function integrates factors such as demand-supply satisfaction, economic considerations, and the CO2 impact of proposed actions.

\subsubsection{Design of the Complementary Hydrological Model}
\label{sub: complementary}

\begin{figure}[ht]
    \centering
    \includegraphics[width=1\linewidth]{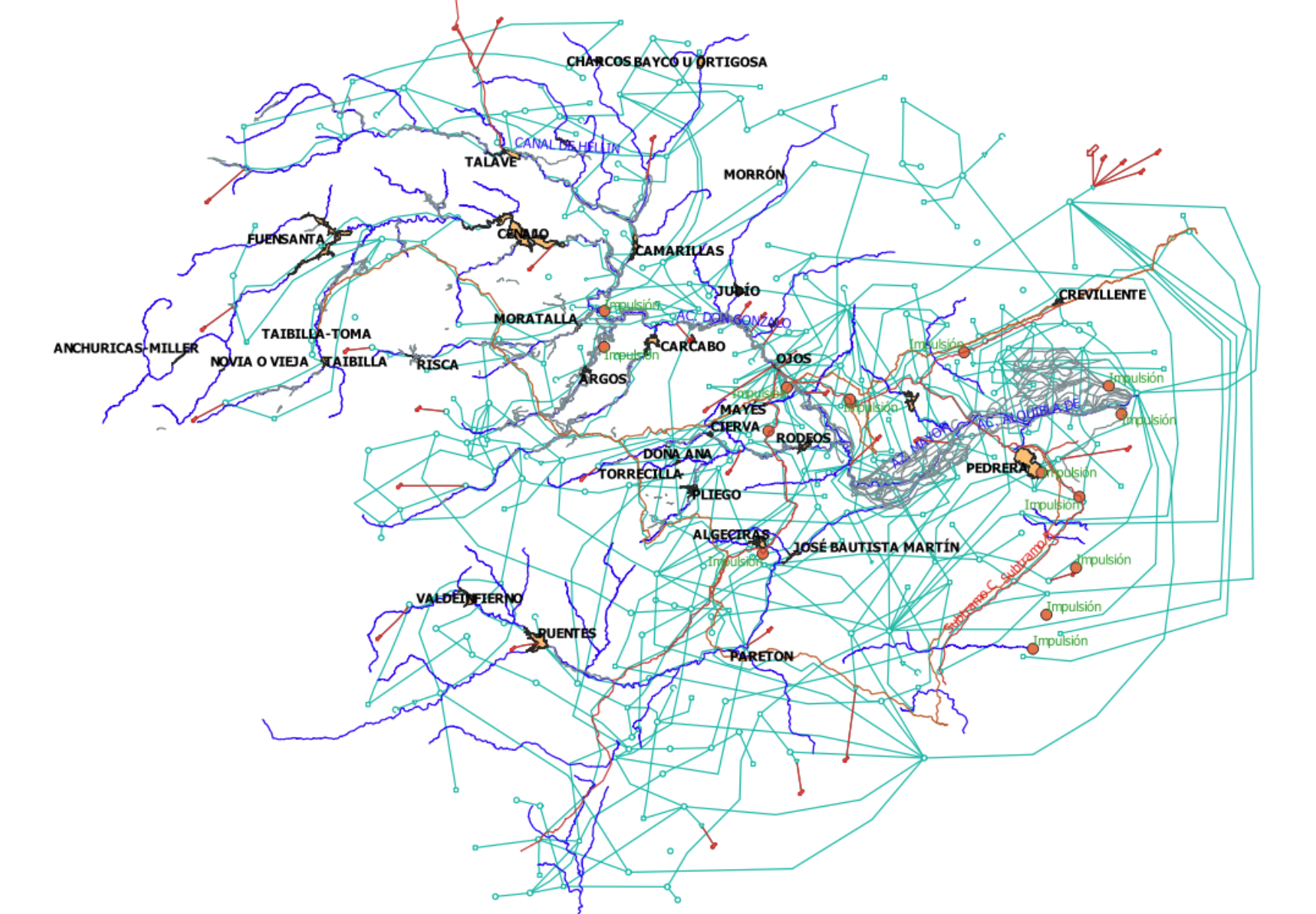}
    \caption{Representation of the CHM showing the interconnected components.}
    \label{fig: CHM}
\end{figure}

The Complementary Hydrological Model (CHM) is developed for comprehensive basin management, incorporating interconnected components, water bodies, demand units and dynamic interactions (refer to Figure \ref{fig: CHM}).

The model is developed using an open-source, advanced network resource allocation library developed by \cite{TOMLINSON2020104635}. This system employs, inter alia, nodes, links, and storage, modelling elements, thus facilitating hydrological modelling and simulation over daily or monthly periods, up to a year, to meet specified requirements.

Data feeding CHM's parameters, as well as water resource and demand status, are sourced from the previous steps of the pipeline: resource prediction (\ref{sub: resource_pred}) and demand estimation (\ref{sub: demand_estim}). 

Overall, the CHM is a versatile, water resource management tool, for flow simulation, thus offering advanced decision-making capabilities. It manages water resources based on end-user operating rules, covering priority of use, rights, guarantees, ecological flows, reservoir operation standards, groundwater limits, aqueduct regulations, and water quality. These rules provide a structured framework for sustainable and equitable water management, optimizing resource allocation and utilization within the CHM framework.

\subsubsection{Cost Function}
\label{sub: costfunc}

The multi-objective cost function driving the optimal resources-to-demands water management in the approach integrates three primary objectives: demand-supply satisfaction, hydro-economic considerations, and CO2 impact assessment (refer to Equation \ref{eq: cost_function}).

\textbf{Demand-supply satisfaction} is a critical metric which can be used to avoid or minimize water deficits within demand units. It compares water income to unit demands, considering the order of preference, ecological flows, and usage rights during water distribution. In the equation the term $\left( \frac{D_i - S_i}{D_i} \right)^2$ represents the deficit metric for each demand unit $i$, emphasizing minimizing water deficits. This metric computes the deficit proportionally to the demand, allowing the optimizer to give equal importance to each unit instead of weighting the units with higher demands. Squaring this proportional deficit enables the algorithm to consider the deficit or surplus equally.

In the context of Agricultural demand areas (ADUs), which represent the principal portion of water demand (87\%), demand estimation is based on agronomic models that provide short- and long-term predictions. Urban demand forecasting relies on empirical curves derived from studies carried out by the \cite{CHS_anejo3}, offering demand projections for each UDU in the forthcoming decade. These projections are defined to consider seasonal variations, ensuring accuracy in estimations, particularly in scenarios marked by elevated water demand, such as during the summer tourism period. Meanwhile, IDUs and SDUs use practically annual raw demand estimations since the CHS considers consistent monthly distribution. The wetland demand (WDU) is assumed to be a constraint, rather than a demand-supply management variable. 

\begin{figure}[ht]
    \centering
    \includegraphics[width=0.85\linewidth]{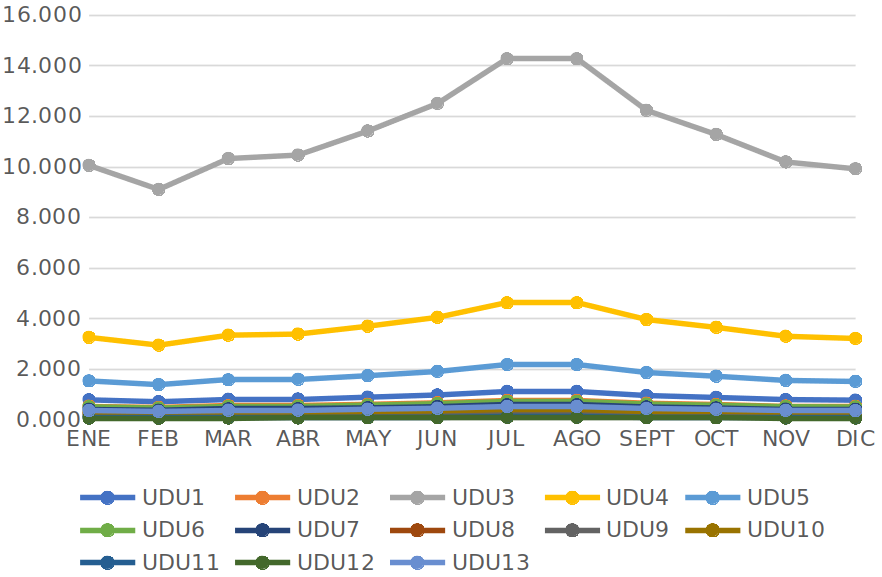}
    \caption{Representation of UDU's demand curve in the year 2021.}
    \label{fig: udu_example}
\end{figure}

The final aim of a $demand-supply$ $ satisfaction$-based water management criterion is the optimal demand-supply, that is, minimizing water deficits while prioritizing the following order: (1) UDU, (2) ADU, (3) IDU, and (4) SDU.

\textbf{CO2 impact} is evaluated by assessing the CO2 emission and absorption factors associated with the activities of demand units, and the environmental effects of the generation water for each source (refer to Table \ref{tab: co2_source}); the latter is determined by considering the energy density ($kWh/m^3$) required for generating each water source, thus applying basin's electric power emission factor ($0.354 kg \ CO2/kWh$), according to \cite{ECOINVENT}, to finally derive the CO2 impact ($kg \ CO2/m^3$). 

In the equation \ref{eq: cost_function}, the $\text{impact}_{CO2}(S_i)$ captures the CO2 emissions related to generating and supplying water to unit $i$.

\begin{table}[htbp]
  \centering
  \caption{CO2 impact corresponding to the energy density associated with the generation of each water source.}
  \label{tab: co2_source}
    \begin{tabular}{rrr}
    
      &  Energy Density & CO2 emission \\
     \toprule
     Surface water & 0.06 & 0.0212 \\
     Groundwater & 0.9 & 0.318 \\
     Desalinization & 4.32 & 1.529\\
     Recycling & 0.78 & 0.276\\
     Transfer & 1.21 & 0.428\\
    \midrule
    \end{tabular}
\end{table}

Assessing CO2 emissions linked to water usage is complex. For agricultural activities, the analytical findings of \cite{martin2020balance} are followed, which objectively measure the role of irrigable areas by estimating greenhouse gas emissions, CO2 absorption from cultivated crops, and overall carbon footprint. Urban demand units (UDUs) are assigned a CO2 emission value of $0.017432137 kg \ CO2/m^3$ \citep{santos2016calculo}. In contrast, industrial emissions vary by economic activity type \citep{CHS_2022}, for instance, the wood industry has an emission of $48.26 kg \ \text{CO2}/m^3$. Additionally, SDUs are attributed an average emission value of $0.6926 kg \ CO2/m^3$ \citep{guairacaja2021estimacion}.

\textbf{Economic Impact} is evaluated by considering the benefits derived from the water demand activities and the costs associated with the generation of water for each source (refer to Table \ref{tab: economic_source}) following the study of the Segura Hydrographic Basin confederation \citep{CHS}. In the equation, the term $\text{impact}_{\text{economic}}(S_i)$ accounts for the economic cost and benefits associated with generating and distributing water to meet the supply $S_i$. Since the objective is to maximize the economic impact, the variable is expressed as a negative value in the cost function.

\begin{table}[htbp]
  \centering
  \caption{Economic cost associated with the water generation of each source \citep{CHS}.}
  \label{tab: economic_source}
    \begin{tabular}{rr}
    
      &  Cost (\euro$/m^3$) \\
     \toprule
     Surface water & 0.003 \\
     Groundwater & 0.25 \\
     Desalinization & 0.6 \\
     Recycling &  0.0 \\
     Transfer &  0.15 \\
    \midrule
    \end{tabular}
\end{table}

For ADUs, demand curves are applied to correlate irrigation water volume with the potential benefits of each crop. Thus, when the decision-making system allocates supply to ADUs, the curve yields the corresponding benefits. IDUs entail creating specific curves for each activity type by determining the ratio between water usage and total benefits based on historical records. Evaluating the economic benefit of UDUs is intricate due to the absence of a defined economic activity objective; thus, benefit measurement proposals are derived from research of \cite{crespo2022integrating}. Lastly, SDU demand benefits, based on the study by \cite{diaz2007competing}, are assigned a value of $12.66\ \text{\euro}/m^3$.

\begin{equation}
\begin{aligned}
   \label{eq: cost_function}
    \text{Minimize:} \quad C &= w_1 \cdot \sum_{i} \left( \frac{D_i - S_i}{D_i} \right)^2 - w_2 \cdot \sum_{i} \left( \text{impact}_{\text{economic}}(S_i) \right) + w_3 \cdot \sum_{i} \left( \text{impact}_{CO2}(S_i) \right) \\
    \text{Subject to:} \quad & \sum_{i} S_i \leq \text{Total Supply} \\
    & S_i \leq D_i \quad \forall i \\
    & \text{Ecological flow constraints} \\
    & \text{Legal constraints on water use rights and transfer regulations} \\
    \text{Where:} \\ 
    & D_i = \text{Demand of unit } i \\
    &w_1, w_2, w_3 = \text{Weights assigned to each objective component} \\
    \text{Decision Variables:} \\ 
    & S_i = \text{Supply allocated to unit } i \\
\end{aligned}
\end{equation}

\subsubsection{Optimization Algorithm}
\label{sub: Algorithm}

The decision-making system employs Mixed-Integer Linear Programming (LP), a mathematical optimization technique, in conjunction with the complementary model to derive decision outcomes based on a cost function.

In the LP model, the decision variables correspond to the flows defined within the hydrological network of the complementary model (refer to subsection \ref{sub: complementary}). These variables are subject to the following constraints:

\begin{itemize}
    \item \textbf{Ecological Flows}: this is the minimum flow of water required to sustain ecological integrity within the basin (\cite{CHS_anejo5}. The volume of ecological flow varies based on the condition, with predefined minimum thresholds for normal and drought scenarios. This ecological flow includes the environmental demand for wetland maintenance (WDU).

    \item \textbf{Minimum Water Resource Levels}: These restrictions define the minimum levels of water resources, particularly for storage bodies. These constraints regulate the maximum extractable water in a period.

    \item \textbf{Water Rights}: these constraints delimit the legal usage rights of each water source across the entire basin. Unlike irrigation rights in ADUs, this encompasses a broader scope and is applicable basin-wide.
\end{itemize}

Integrating the constraints ensures a solution following ecological preservation mandates, resource sustainability, and regulatory frameworks.

\section{Results and Discussion}
\label{sec: Results}

This section presents the results of real water management planning in the Segura Hydrographic Basin. The analysis includes a long-term optimization covering six months, from April 2, 2024, to September 29, 2024. The objective function emphasizes minimizing the deficit ($w_1=0.6$), over the environmental ($w_2=0.3$) and economic impact ($w_3=0.1$).

\begin{figure}[ht]
    \centering
    \includegraphics[width=\linewidth]{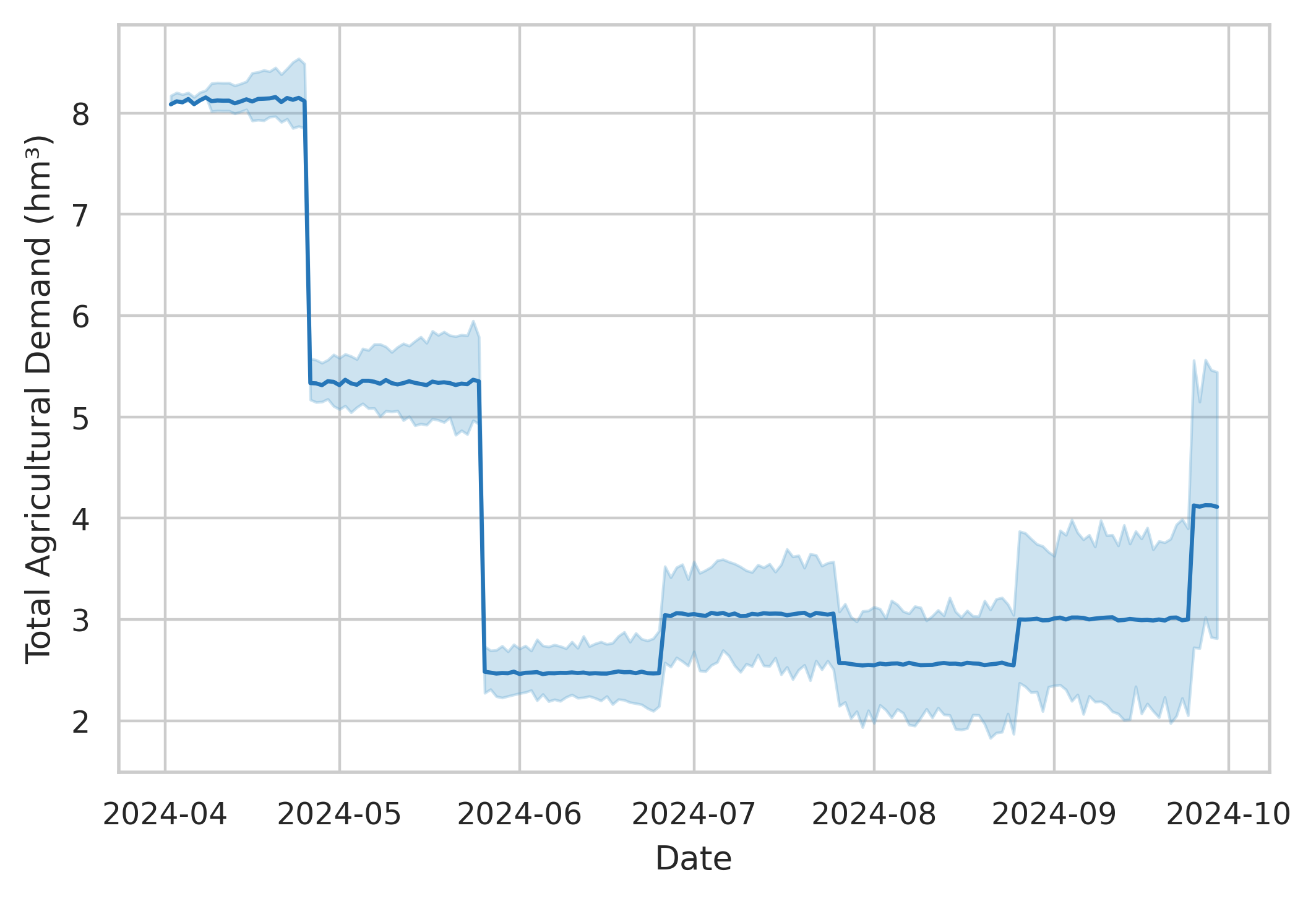}
    \caption{Evolution of the total estimated agricultural demand ($hm^3$) corresponding to the period of the exemplary use case. The shadows represent the uncertainty of the estimation.}
    \label{fig: uncertainty}
\end{figure}

During the above period, a supply of approximately $641.92\ hm^3$ of water is assigned to meet an estimated demand of $710.24\ hm^3$ (refer to Table \ref{tab: results_demand_per_unit}). The decision-making system projects an approximate deficit of $69\ hm^3$ of water, representing 9.7\% of the total demand. It must be noted that the estimated demand is subject to variations and uncertainties due to the prediction models. Consequently, the longer the prediction period, the larger the uncertainty introduced into the algorithm (refer to Figure \ref{fig: uncertainty}). Hence, the deficit obtained in this decision-making process will vary as the planning period progresses.

\begin{table}[htbp]
  \centering
  \caption{Resource usage resulted in the decision-making example case.}
  \label{tab: results_resources}
    \begin{tabular}{rr}
    
       & Usage ($hm^{3}$) \\
     \toprule
     Surface water & 383.18 \\
     Groundwater & 22.78  \\
     Desalinization & 28.66 \\
     Recycling & 71.24 \\
     Transfer  & 136.06 \\
    \midrule
     & 641.92 $hm^{3}$\\
    \end{tabular}
\end{table}

\begin{figure}[ht]
    \centering
    \includegraphics[width=\linewidth]{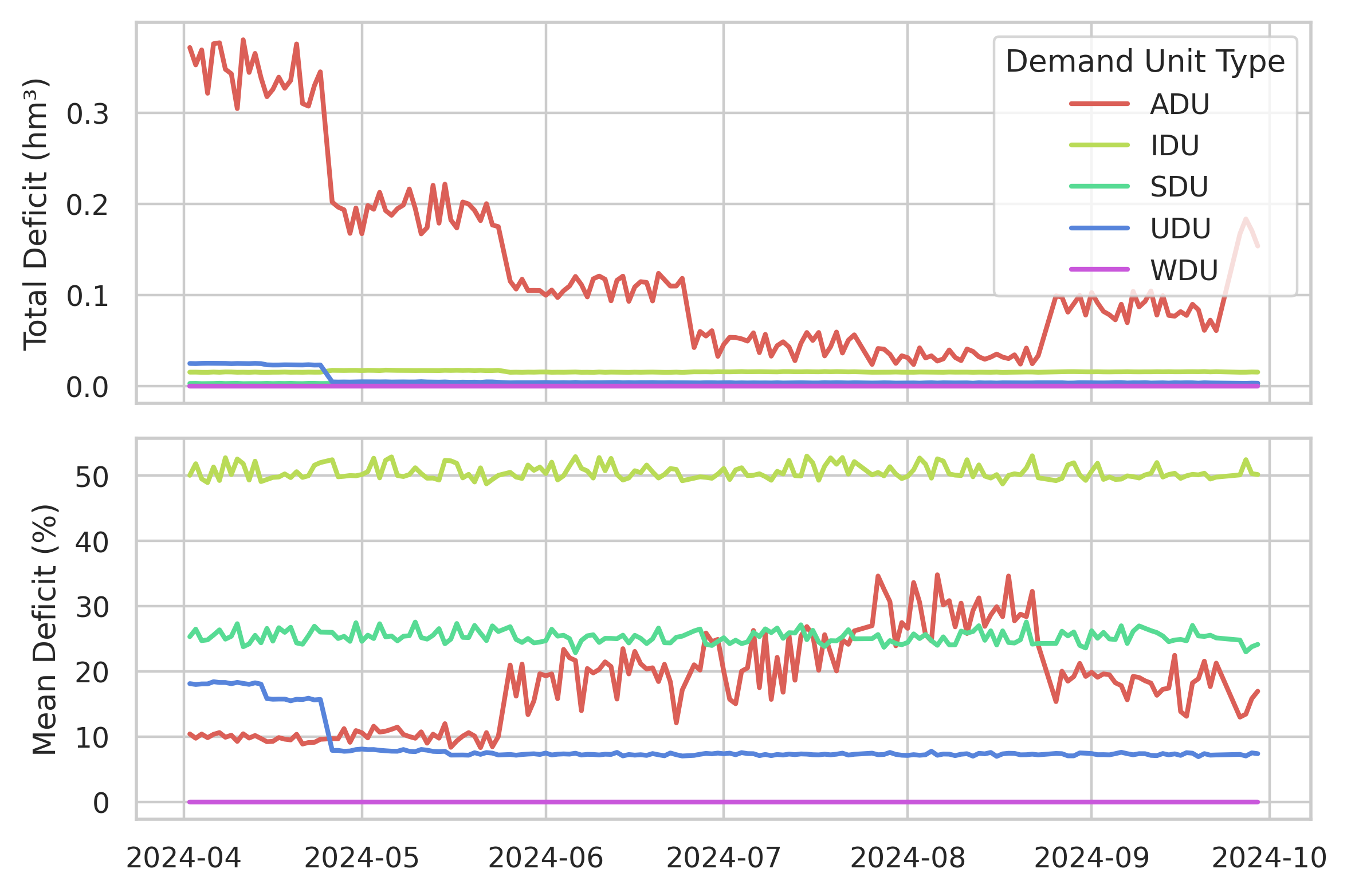}
    \caption{Timeline of the daily evolution of the total water deficit ($hm^3$) and average percentual deficit for each water demand type.}
    \label{fig: deficit_plot}
\end{figure}

Regarding the obtained results, the effect of the weighting of the different objectives in the cost function can be clearly seen in this example solution. The economic impact has a low weight ($w_3=0.1$), causing the deficit to primarily affect the units that generate the highest economic cost, namely the agricultural and industrial demands while prioritizing the urban demand fulfilment. Figure \ref{fig: deficit_plot} illustrates the daily deficit for each type of demand, by aggregating the deficit of individual demand units for calculating a total deficit per demand type and average percentual deficit per demand unit. 

The water deficit impacts agricultural demand the most since it represents the most demanding activity. However, due to the lower priority of the economic impact in the optimization function, industrial demand units experience a higher proportional deficit, nearing 50\% of their demand. In this sense, although both agricultural and industrial activities generate substantial economic impacts, industrial activities contribute more to CO2 emissions compared to agricultural activities. Consequently, the optimization algorithm prioritizes agricultural demand over industrial demand.

Moreover, the average percentage deficit trend indicates that urban demand units (UDUs) are prioritized over agricultural, industrial, and service demands. However, urban supply can be affected, as seen in the first month of the timeline, due to tougher quality and source restrictions.  As indicated above, the water demand for wetlands is introduced as a constraint in the system, thus resulting in a corresponding net zero deficit.

\begin{table*}[htbp]
  \centering
  \caption{Water supply distribution, CO2 and economic impact for each demand type from April 2, 2024, to September 29, 2024, obtained in the example case using the proposed decision-making pipeline.}
  \label{tab: results_demand_per_unit}
    \begin{tabular}{rrrrr}
    
         & Supply ($hm^{3}$) &  Pred. Demand ($hm^{3}$) & Emissions (t CO2eq) & Economic (mill. \euro) \\
     \toprule
     Agricultural & 506.43 & 570.11 & 526,177 & 23,414.09 \\
     Urban & 114.63 & 116.02 &  2,342 &  1,720.86 \\
     Industrial & 1.92 & 4.72 & 1,178,296 & 12,238.15 \\
     Service & 5.19 & 5.63 &  4,236 & 76.95 \\
     Wetland  & 13.75 & 13.75 &   -5,583 & 0\\
    \midrule
     & 641.92& 710.24& 1,705,468 & 37,450.05\\
    \end{tabular}
\end{table*}

In terms of environmental impact, Table \ref{tab: results_resources} shows that surface waters are the most exploited, with 60\% of total use since surface water bodies have the lowest economic and environmental cost (refer to Tables \ref{tab: co2_source} \& \ref{tab: economic_source}). Conversely, desalinated waters have the highest environmental impact and are among the least utilized, with 4.4\% of total use.

Transfer waters are the second most utilized despite having a considerable economic and environmental impact. As shown in Table \ref{tab: udu_waters}, urban demand mainly consumes surface and transfer waters due to quality restrictions that prevent these demands from using recycled and desalinated waters. Given that minimizing the deficit is significantly more important than economic or environmental impact according to the above weighting, the system prioritizes this decision at the expense of the absolute cost.

\begin{table}[htbp]
  \centering
  \caption{Total supply ($hm^{3}$) and resource usage distribution in the example case for each Urban Demand Unit.}
  \label{tab: udu_waters}
    \begin{tabular}{lllll}
    \toprule
     Unit &  Supply &  Surface water &  Groundwater &   Transfer \\
    \midrule
    UDU03 &  26.033 &       26.033 &        0.0 &               0.0 \\
    UDU04 &  19.461 &       19.461 &        0.0 &               0.0 \\
    UDU05 &  19.412 &       19.412 &        0.0 &              0.0 \\
    UDU07 &  16.095 &        0.0 &        0.0 &             16.095 \\
    UDU01 &   9.541 &        9.541 &        0.0 &           0.0 \\
    UDU02 &   9.403 &        9.403 &        0.0 &            0.0 \\
    UDU06 &   5.894 &        0.0 &        0.0 &             5.894 \\
    UDU14 &   2.452 &        2.452 &        0.0 &           0.0 \\
    UDU08 &   2.154 &        2.154 &        0.0 &       0.0 \\
    UDU09 &   1.609 &        1.609 &        0.0 &          0.0 \\
    UDU11 &   1.128 &        0.882 &        0.246 &        0.0 \\
    UDU12 &   0.624 &        0.624 &        0.0 &          0.0 \\
    UDU13 &   0.475 &        0.475 &        0.0 &         0.0 \\
    UDU10 &   0.350 &        0.0 &        0.350 &         0.0 \\
    \bottomrule
    \end{tabular}
    
\end{table}

It can be seen that groundwater usage is almost residual, which is currently a positive result, explained by continuous over-exploitation in the past. This occurs since groundwater costs are significantly higher than surface water sources (refer to Table \ref{tab: economic_source}), an aspect that has not been considered in the planning processes followed in the past.

%Thus, this real water planning example demonstrates the utility of the proposed decision-making system, which is based on water availability estimates and demand predictions given by AI models, the simulation of a complementary hydrological model, and a multiobjective cost function. The weights assigned to each objective of the cost function play a crucial role, guiding the optimization system to minimize costs and water deficits, by simulating different scenarios and prioritizing the optimum ones. Additionally, as observed, the system considers and meets constraints and limitations, such as water quality compliance criteria, minimum ecological flows and, wetland water demand.

The results demonstrate that the proposed methodology significantly advances environmental resource management by transitioning from traditional static models to a dynamic, AI-driven approach. As shown by the Segura Hydrographic Basin case, conventional IWRM systems that rely on historical data and static simulations are insufficient for addressing the complexities of modern hydrological challenges. In contrast, our approach integrates advanced modelling with real-time data inputs, weather forecasting, and distributed hydrological simulation to create a forward-looking decision-making system. This system continuously adapts to current and predicted environmental conditions, leading to more accurate water availability and demand predictions. Consequently, as shown in figure \ref{fig: deficit_plot}, this allows for more efficient resource allocation, thus reducing wastage and improving sustainability.

Incorporating artificial intelligence models into the decision-making process also introduces adaptability and precision unreachable using conventional methods. As indicated by the results in table \ref{tab: results_demand_per_unit}, the system optimizes resource distribution while considering environmental indicators such as CO2 emissions and economic impacts. This guarantees that decisions are made to minimize environmental impact and advance long-term sustainability rather than just satisfying short-term water demands. For instance, optimizing irrigation schedules in agriculture demonstrates a significant reduction in water usage and associated CO2 emissions, obtaining both environmental and economic benefits. These findings highlight significant improvements offered by this methodology compared with traditional approaches, which often fail to account for the dynamic nature of environmental systems and the broader impacts of water management decisions.

Moreover, the generalizability of this approach is noticeable since it relies on globally accessible weather forecasts and remote sensing data, enabling its application to various geographical regions beyond the Segura Hydrographic Basin. However, the physical modelling components, specifically the Distributed Hydrological Model (DHM) and Complementary Hydrological Model (CHM), must be modified to reflect the characteristics of different basins. This adaptation implies adjusting the models for local hydrological features, such as topography and existing water infrastructure. Once these adjustments are performed the approach can be executed in other basins, providing a robust and scalable solution for water resource management in diverse environments.

\subsection{Integration}
\label{sec: Integration}

The proposed pipeline has been developed in the context of a pre-commercial public procurement tendering framework and undertaken in close collaboration with government institutions and the Segura Hydrographic Confederation (CHS). Consequently, this collaboration has facilitated the integration of the proposed approach into the Integrated Water Resources Management framework of the Rivera Basin. This section briefly overviews the integration process of the decision-making system into an operational scenario.

\begin{figure}[ht]
    \centering
    \includegraphics[width=\linewidth]{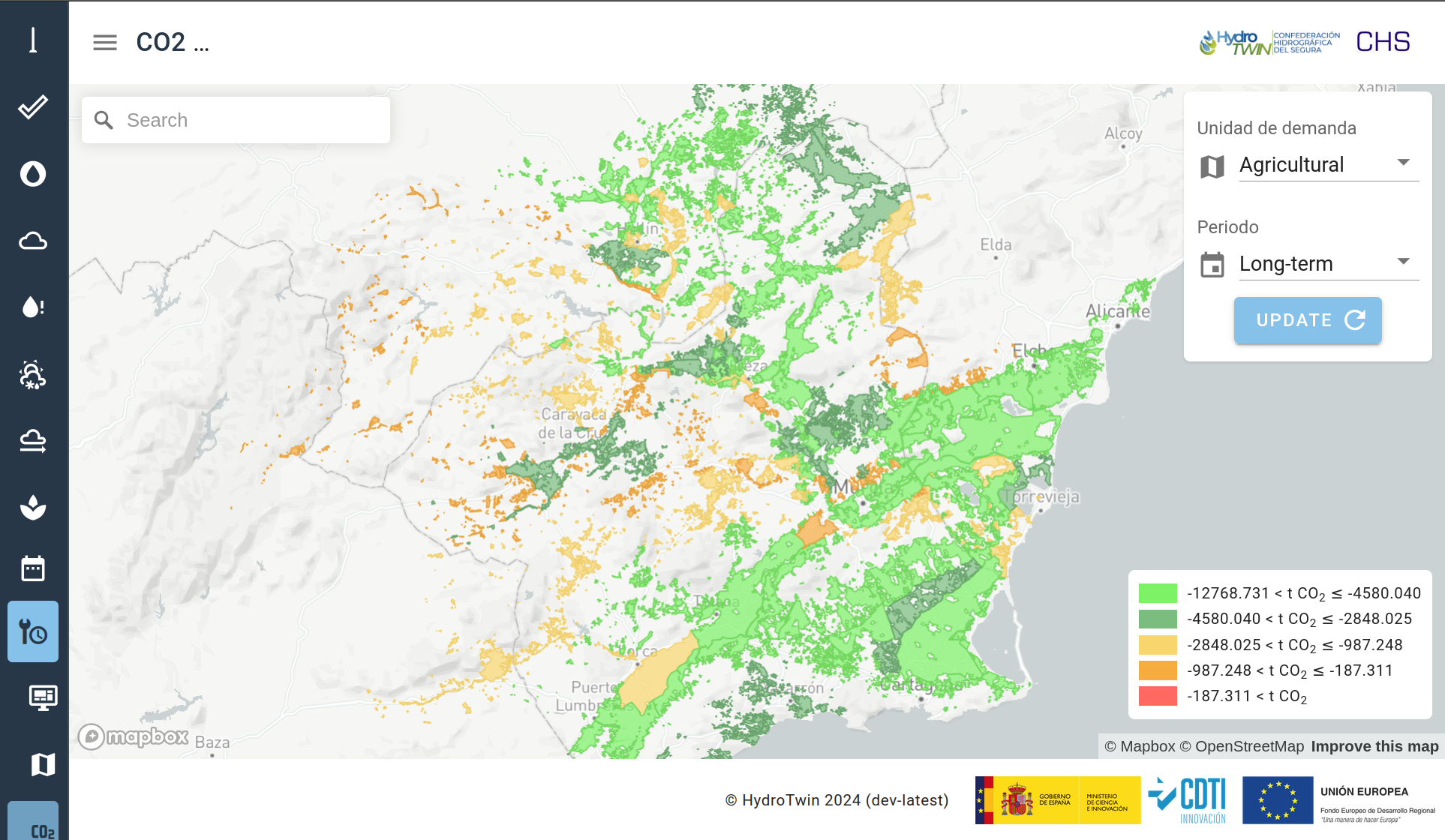}
    \caption{Operational example of the decision-making framework integrated into a Web application. The visualization shows the accumulated CO2 emissions per ADU in the short-term prediction period.}
    \label{fig: co2_interface}
\end{figure}

The proposed framework operates as a semi-automated system, wherein specific pipeline components are periodically executed, while others require user interaction and/or configuration. Consequently, the developed solution is integrated into a software architecture comprising a secured, user-friendly Web application (refer to Figure \ref{fig: co2_interface}).

The implemented architecture is integrated within the operational infrastructures of the CHS, thus enabling interaction with its real-time monitoring systems, such as dam monitoring. Furthermore, historical records are extracted from the CHS and stored in a high-speed read-and-write InfluxDB database, specifically designed for time series management. Using a local replica of the records enhances the efficiency of data reading, processing, and visualization.

The weather forecasting and hydrological simulations in the DHM are executed daily, thus reducing uncertainty in the planning scenario, especially in the first days. These demanding algorithms are executed in distributed computing servers. Also, the generated geospatial predictions are stored in a common data lake for efficient data management.

Similarly, the periodic predictions of water resources and estimation of demands enable the decision-making system to optimise water planning with reduced uncertainty. Consequently, end-users perform management decisions based on more accurate estimations. Interaction with the Web application allows experts to generate, analyze, and validate new water management solutions, incorporating the desired water distribution priorities ($w_1,w_2,w_3$) in the cost function (refer to Section \ref{sub: costfunc}). 

Additionally, the developed Web application provides users with tools for exploring and analyzing the data employed in the decision-making process, including the real-time monitoring of data, user-demanded forecasting of operational variables and, crop irrigation information, such as historical irrigation, estimated and predicted water demand and, irrigation proposal, through a user-friendly, interactive map service.

\section{Conclusions}
\label{sec: conclusions}
This paper presents an Integrated Water Resources Management pipeline for decision-making in the Segura Hydrographic Basin. The research work introduces a novel approach for predicting available water resources and demands by leveraging advanced techniques, such as the integration of remote sensing data and deep learning models for agricultural crop growth forecasting. The proposed methodology offers a comprehensive solution to address the optimal management of water.

The basin is known for its complex river systems, as documented by \cite{grindlay2011atomic}, and presents critical challenges such as limited water resources, governance complexities, pollution incidents, and prolonged droughts. Given the significant agricultural activities, accurate water demand estimation is contingent upon measuring crop dynamics and production requirements.
 
The proposed methodology integrates real-time monitoring systems within the basin with advanced weather forecasting and physical simulation models to predict forthcoming water resource availability (refer to Section \ref{sub: resource_pred}). The fluctuation of irrigation needs in agricultural activities is estimated using high-resolution irrigation demand maps generated by processing remote sensing data using cutting-edge AI techniques (refer to Section \ref{sub: demand_estim}). These forecasts are used in a decision-making algorithm to optimize economic and environmental outcomes, prioritizing demand-supply balance, ecological flows, and historical water use rights of basin resources (refer to Section \ref{sub: Decision-making}). The proposed platform is finally integrated into CHS's operational framework, through close collaboration with the Segura Hydrographic Confederation. Integrating an interactive Web application facilitates obtaining valuable feedback on performance aspects and enhances the understanding of expert-user interaction dynamics. Thus, with the development of the above water management tool, it can be said that this research work has opened a window towards the consideration of cutting-end digital tools for river basin water management, considering physical models and river basin simulators, operational predictive models based on AI, and models for estimating the economic impact and environmental impact of the various uses of water that, to date, have not been taken into account.

Future work should focus on improving the prediction of agricultural demand and reducing uncertainty, as this is the critical factor of optimal water management in the basin. Incorporating additional data sources, such as soil moisture sensors and localized weather data, can enhance the accuracy of crop growth forecasting. Collaborating with agricultural stakeholders and conducting comprehensive field measurements will also improve the robustness and precision of predictions. Refining deep learning algorithms with advanced techniques like Generative Adversarial Networks (GANs) could allow the simulation of more diverse scenarios. Finally, incorporating fine-tuning, human-in-the-loop-based advanced algorithms for more accurately validating economic, environmental, and social considerations will be valuable.

Overall, the work introduces a solution for optimum water management based on artificial intelligence techniques. While the technical intricacies and mathematical validations of the techniques are beyond the scope of this paper, they represent a crucial direction for future exploration and refinement. This decision-making tool enables the simulation of diverse and dynamic scenarios endowing decision-makers with a higher accuracy in their estimations.

% Acknowledgements 
\section*{Acknowledgment}
\label{section:acknowledgment}

This work was supported by project GEMELO DIGITAL CHS (IN3210/2021), financed by the \textit{CDTI-CPI-2020} program of the \textit{Centro para el Desarrollo Tecnologico y la Innovacion} (CDTI) and run in collaboration with \textit{Confederación Hidrográfica del Segura} (CHS), Grusamar, DHI, Meteobit and Vicomtech Foundation.

%The authors would like to extend sincere gratitude to Radoslaw Guzinski for providing the algorithm used in this research. His contribution was invaluable to the successful application of the model in this work.
\bibliography{sample}

\end{document}